\documentclass[10pt,twocolumn,letterpaper]{article}
\usepackage[pagenumbers]{iccv}
\usepackage{microtype}
\usepackage{placeins}
\definecolor{iccvblue}{rgb}{0.21,0.49,0.74}
\usepackage[breaklinks,colorlinks,allcolors=iccvblue]{hyperref}
\usepackage[accsupp]{axessibility}
\def\paperID{7}
\def\confName{ICCV}
\def\confYear{2025}
\title{ManzaiSet: A Multimodal Dataset of\\Viewer Responses to Japanese Manzai Comedy}
\author{Kazuki Kawamura$^{1,2}$ \quad Kengo Nakai$^{3}$ \quad Jun Rekimoto$^{1,2}$\\[2pt]
$^{1}$Sony CSL Kyoto \quad $^{2}$The University of Tokyo\\
$^{3}$Yoshimoto Kogyo Holdings Co., Ltd.\\
{\tt\small kwmr@acm.org \quad nakai.kengo@yoshimoto.co.jp \quad rekimoto@acm.org}}
\hypersetup{pdftitle={ManzaiSet: A Multimodal Dataset of Viewer Responses to Japanese Manzai Comedy},pdfauthor={Kazuki Kawamura, Kengo Nakai, Jun Rekimoto}}
\begin{document}
\maketitle
\begin{abstract}
We introduce ManzaiSet, a multimodal dataset of individual viewer responses to Japanese manzai comedy.
The corpus contains 2,301 viewer--performance recordings from 241 native Japanese speakers, totaling 191.8 hours of facial video and microphone audio.
Participants watched up to 10 shared professional performances in randomized order at home and rated each performance on a 0--100 scale.
Exploratory analyses yielded three findings. First, viewers differed in both their average ratings and how much their ratings varied across performances. Viewers who rated one random half of the performances highly also tended to rate the other half highly within the same viewing sequence. Second, performances watched later in the sequence received higher ratings on average. Third, after multiple-comparison correction, no statistically significant differences in viewer responses were found between humor categories. ManzaiSet supports the study of individual differences by linking shared Japanese comedy performances to viewer ratings, viewing order, and audiovisual responses.
\end{abstract}

\begin{figure}[t]
\centering
\includegraphics[width=\columnwidth]{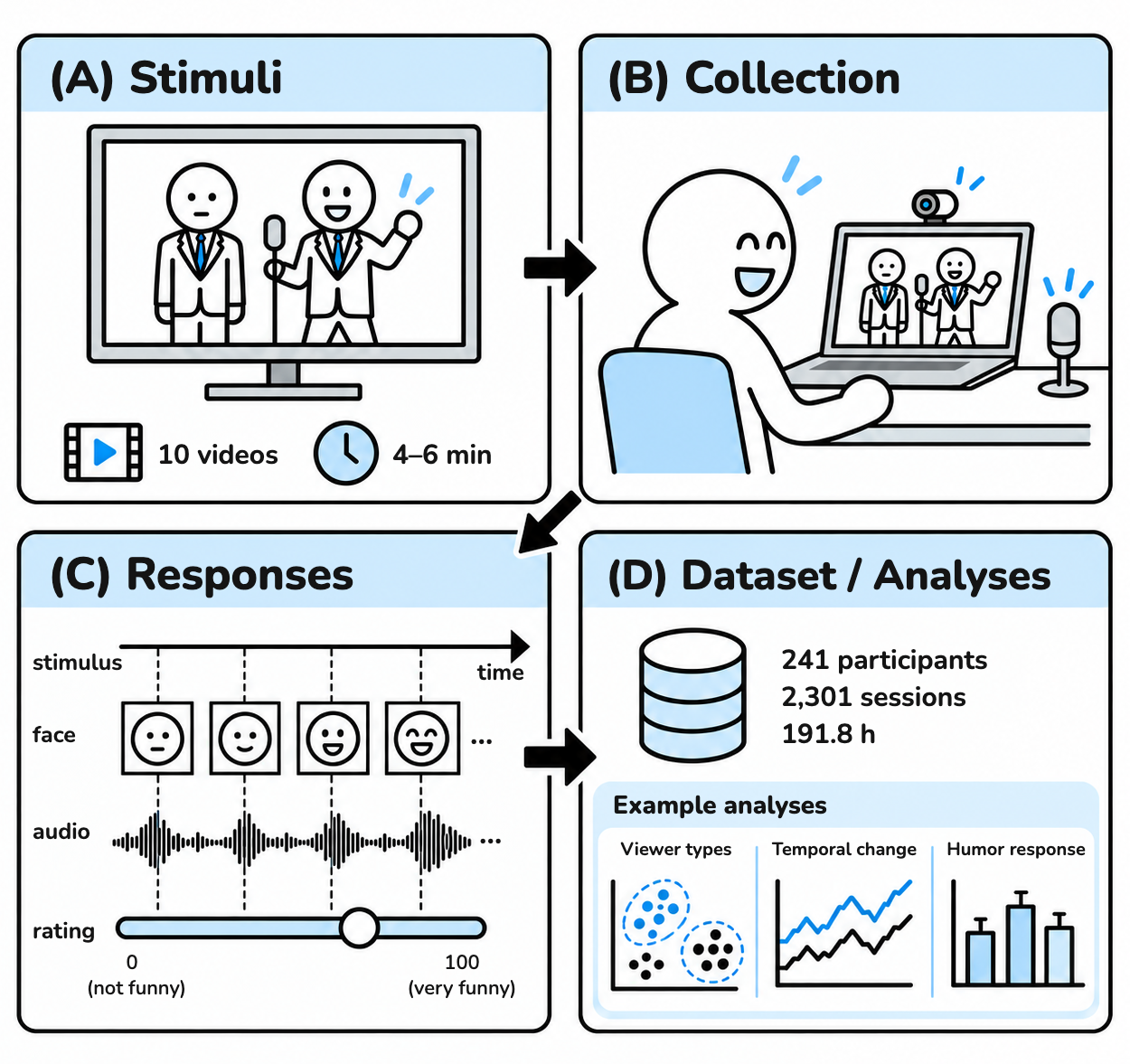}
\caption{Overview of ManzaiSet. Participants watch professional manzai performances at home while facial video and audio are synchronously recorded, followed by a subjective rating after each video. The resulting dataset contains 2,301 viewing sessions from 241 participants (191.8 hours) and enables analyses of individual viewer differences, temporal dynamics, and humor responses. The rating panel schematically illustrates the 0--100 overall evaluation.}
\label{fig:setup}
\end{figure}

\section{Introduction}
Viewers can respond differently to the same entertainment content.
Studying this variation requires a connection between the content presented, each viewer's observable responses, and that viewer's own evaluation.
For comedy, this connection spans several timescales: vocal and facial activity unfolds during a performance, a rating summarizes the performance, and successive ratings describe a viewing session.
A dataset that preserves these relationships supports questions about variation between viewers, variation between performances for the same viewer, and changes across viewing positions.

Manzai is a Japanese form of two-person comedy organized around exchanges between the \emph{boke}, who introduces incongruity, and the \emph{tsukkomi}, who responds to or corrects it~\cite{stocker2006manzai,tsutsumi2011boke}.
Repeated exchanges and recurring material provide a temporally structured setting for recording responses to professional entertainment.
Language, delivery, and familiarity with comic conventions are part of this viewing context.
The shared performance timeline makes it possible to connect responses from separately recorded viewers to the same content.

ManzaiSet combines 10 professional manzai performances with 2,301 viewer--performance recordings from 241 native Japanese speakers.
Each participant watched multiple performances in randomized order and provided a rating after each one.
The 191.8 hours represent the summed duration of participant recordings across the shared stimulus set.
The repeated unit is a performance within an assigned viewing sequence.
This structure supports comparisons among viewers exposed to common content and among performances viewed by the same person.

Shared-stimulus recording has a substantial history in affective computing, including DEAP, MAHNOB-HCI, and SEWA~\cite{Koelstra2012DEAP,soleymani2012multimodal,kossaifi2021sewa}.
Individual audiovisual laughter recordings are also available in MAHNOB Laughter~\cite{petridis2013mahnob_laughter}.
ManzaiSet extends this line of work through its sustained professional manzai stimuli, repeated performance-level observations, recording volume, and linkage of individual signals with ratings and viewing order.

We examine three aspects of viewer response: individual rating patterns, changes across viewing positions, and differences between humor categories.

\section{Related Work}
\subsection{Stimulus-Linked and Naturalistic Affect Corpora}
DEAP records physiological responses from 32 participants watching music videos, with frontal face video available for 22 participants~\cite{Koelstra2012DEAP}.
MAHNOB-HCI links audiovisual and physiological recordings to responses elicited by multimedia stimuli~\cite{soleymani2012multimodal}.
AMIGOS includes individual and group viewing settings together with affect, personality, and mood information~\cite{miranda2018amigos}.
LIRIS-ACCEDE provides movie excerpts and annotations of affective content~\cite{baveye2015lirisaccede}.

SEWA records 398 participants from six cultural groups using their own computers, including responses to shared advertisements and subsequent video-chat discussions~\cite{kossaifi2021sewa}.
Its common advertisements and remote recording environment are closely related to the collection design used here.
ManzaiSet concentrates on 10 professional comedy performances and repeated performance-level ratings within a comedy-viewing sequence.
RECOLA captures remote collaborative interactions and provides continuous affect annotations~\cite{ringeval2013recola}.
The interactive tasks in these corpora and the individual entertainment-viewing task in ManzaiSet provide complementary contexts for studying behavior.

AffectNet comprises Internet facial images with expression and valence/arousal annotations~\cite{mollahosseini2019affectnet}.
Aff-Wild2 provides videos for expression, action-unit, and affect analysis~\cite{kollias2019affwild2}.
CAER combines face and scene information for contextual emotion recognition~\cite{lee2019caer}.
Their emphasis is recognition from observed images or videos.
ManzaiSet additionally links separate viewers to a known shared stimulus and a performance-level rating.

BP4D+ combines spontaneous behavior, multimodal recording, self-report, and facial-action annotation~\cite{zhang2016bp4dplus}.
4DFAB includes repeated recording sessions over an extended period~\cite{cheng2018fourdfab}.
ManzaiSet records responses to different performances within the same assigned viewing sequence.

\subsection{Humor Content and Individual Responses}
UR-FUNNY pairs multimodal content with preceding discourse context for humor detection~\cite{Hasan2019URFUNNY}.
SMILE studies laughter-related video understanding with language models~\cite{lee2024smile}, while StandUp4AI and multilingual stand-up laughter detection extend content analysis across languages~\cite{barriere2025standup4ai,kuznetsova2024lrec}.

MAHNOB Laughter directly records individuals' faces and voices, including a session in which participants watch funny video clips~\cite{petridis2013mahnob_laughter}.
It contains 22 participants and 3 hours 49 minutes across its recording tasks.
ManzaiSet develops the individual-response approach around a shared set of sustained professional performances, with 241 viewers, performance-level ratings, and viewing-order metadata.
The resulting unit of comparison is a viewer's response to a particular performance or stimulus instance.

\subsection{Manzai and Cultural Context}
Conversation analyses describe the organization of boke--tsukkomi exchanges and manzai-like collaborative humor in Japanese interactions~\cite{tsutsumi2011boke,zawiszova2021manzai}.
Work on Japanese puns also identifies the role of phonological and psychoacoustic similarity~\cite{kawahara2009puns}.
These studies situate the language and interactional structure of the selected stimuli.
ManzaiSet uses automated humor tags to describe instances within the selected performances.

Research on human emotion perception and emotional display rules documents cultural variation in recognition and expression~\cite{elfenbein2002cultural,jack2012pnas,matsumoto2008displayrules}.
ManzaiSet represents the recruited native Japanese-speaking viewers responding at home to a particular set of Japanese performances.

\section{Dataset Construction}

\subsection{Participants and Stimuli}
We recruited 241 compensated native Japanese speakers: 39 women, 201 men, and one participant reporting another gender.
Ages ranged from 19 to 63 years, with a mean of 30.0 and median of 27.0.
For general comedy viewing frequency, 135 participants (56.0\%) selected the maximum response of 5 on a five-point scale.
For manzai-specific viewing, 114 participants (47.3\%) selected 4 on a five-point scale.

The stimulus set comprises 10 professional manzai performances selected from official recordings of \emph{Densetsu no Ichinichi} (A Legendary Day), held at Namba Grand Kagetsu on April 2--3, 2022, for Yoshimoto Kogyo's 110th anniversary.
Each selected performance lasts approximately four to six minutes.
Selection considered video quality, absence of copyrighted background music, and variety in comic material, including verbal and physical humor.
All participants drew from this common set, presented in randomized order.

\begin{table}[t]
\centering\small
\caption{Corpus units and analysis samples. Duration sums recordings across viewers exposed to the shared set of 10 performances.}
\label{tab:dataset}
\begin{tabular}{lr}\toprule
Quantity & Value\\\midrule
Participants & 241\\
Shared performances & 10\\
Viewer--performance recordings & 2,301\\
Recordings per participant (mean) & 9.55\\
Summed recording duration & 191.8 h\\
Facial video & $640\times360$, 25 fps\\
Microphone audio & 48 kHz\\
Performance rating (mean $\pm$ SD) & $82.0\pm16.4$\\
A/B: viewers with $\geq8$ performances & 228\\
C: viewers with complete response data & 207\\
C: stimulus instances / tag assignments & 77 / 131\\\bottomrule
\end{tabular}
\end{table}
\subsection{Procedure and Rating Measure}
Participants used a browser-based application built with Flask on their own computers at home (Fig.~\ref{fig:setup}).
They logged in with a participant identifier, completed demographic and comedy-preference questions, and checked their camera and microphone.
Facial video and microphone audio were recorded during performance playback.
After playback, the application saved the recording and participants provided a performance rating.
Participants could take breaks between performances and resume the viewing sequence later.
The participant instructions also allowed returning to a previous video and replaying it from the start.
The repeated unit in the analyses is the recorded viewer--performance observation within the assigned viewing sequence.

The participant instruction was: ``Before moving to the next video, please rate the video you just watched out of 100.''
The resulting \emph{performance rating} is an overall evaluation on a 0--100 scale.

\subsection{Recording Units and Data Layers}
A \emph{viewing session}, or \emph{viewer--performance recording}, is one participant's recording while watching one performance.
The corpus contains 2,301 such recordings, an average of 9.55 per participant, totaling 191.8 hours (Table~\ref{tab:dataset}).
Facial video was recorded at $640\times360$ pixels and 25 frames/s, with microphone audio at 48 kHz.
Participant, performance, and viewing-position identifiers connect each recording to its rating and place in the viewing sequence.
Event-based analysis uses stimulus onset times and the corresponding response windows within a performance.
Fine-grained temporal analyses depend on playback-to-recording alignment and the timing variability of participants' browsers and devices.

The recorded signals are facial video and microphone audio.
Participant-provided ratings are performance-level observations.
OpenFace 2.0~\cite{baltrusaitis2018openface} supplies derived facial-action and head-pose features.
Automated tags describe candidate humor instances in the stimuli.
Audio processing supplies candidate vocal responses for stimulus-linked summaries.

OpenFace tracking met the criteria \texttt{success=1} and \texttt{confidence>0.9} in 96.3\% of frames.
Participants were encouraged to use headphones to reduce stimulus sound entering the microphone.
Microphone signals can also contain speech, background sounds, and stimulus leakage; these sources affect audio-derived response measures.

\begin{table}[t]
\centering\small
\caption{Cluster-number diagnostics for standardized rating features. The three-cluster solution provides an interpretable summary of this sample.}
\label{tab:diagnostics}
\begin{tabular}{rcc}\toprule
$k$ & Silhouette & Calinski--Harabasz\\\midrule
2 & 0.515 & 148.14\\
3 & 0.523 & 151.87\\
4 & 0.476 & 158.03\\
5 & 0.488 & 149.93\\\bottomrule
\end{tabular}
\end{table}
\begin{table}[t]
\centering\small
\setlength{\tabcolsep}{3pt}
\caption{Rating profiles among 228 viewers. Columns represent high-rating, variable-rating, and lower-rating profiles. Mean, SD, CV, and range are averages of participant-level features. Trends indicate the direction of the average viewing-order slope.}
\label{tab:profiles}
\begin{tabular}{lrrr}\toprule
 & \shortstack{High ratings\\low variability} & \shortstack{Variable ratings\\positive trend} & \shortstack{Lower ratings\\negative trend}\\\midrule
$n$ & 166 & 32 & 30\\
Share (\%) & 72.8 & 14.0 & 13.2\\
Mean & 87.0 & 71.2 & 65.6\\
SD & 4.84 & 17.9 & 13.9\\
CV & 0.058 & 0.263 & 0.218\\
Range & 15.6 & 57.8 & 43.7\\
Trend & Slightly positive & Positive & Negative\\\bottomrule
\end{tabular}
\end{table}
\subsection{Participation and Use Conditions}
The participant information stated that facial video, audio, and questionnaire information would be used for research, managed strictly, and kept from unrestricted public disclosure.
It explained that participation constituted agreement to the stated conditions and that participants could stop the experiment at any time.
The corpus is therefore a restricted research collection.
Access to identifiable recordings must remain consistent with the participant information and the permissions governing their use.
The professional performance videos are separately subject to the rights of their owners.

\section{Exploratory Analyses}
Analyses A and B include the 228 participants who watched at least eight performances (94.6\% of the recruited sample).
The remaining 13 participants contribute to the corpus totals.
Analysis C uses 207 participants with complete response data for the relevant stimulus instances.

\subsection{A: Descriptive Rating Profiles}
\paragraph{Method.}
For each participant, we summarized ratings using six features: mean, across-performance standard deviation (SD), coefficient of variation (CV, SD divided by mean), range, difference between late- and early-half mean ratings, and a linear slope over viewing positions.
We standardized the features and applied $k$-means clustering.
Silhouette and Calinski--Harabasz diagnostics for $k=2$ through $5$ are shown in Table~\ref{tab:diagnostics}.
We selected $k=3$ as an interpretable descriptive summary: it has the highest silhouette score among these solutions, while $k=4$ has the highest Calinski--Harabasz score.

We also correlated participant mean ratings across randomized halves of the same viewing sequence.
This statistic evaluates rank-order consistency of mean ratings among participants within that sequence.

\paragraph{Results.}
Table~\ref{tab:profiles} reports the three descriptive profiles, ordered by mean rating.
The largest profile contains 166 viewers (72.8\%), with high mean ratings and comparatively low across-performance variability.
A second profile contains 32 viewers (14.0\%) with a mean rating of 71.2, greater variability, and a positive viewing-order trend.
The remaining profile contains 30 viewers (13.2\%) with a mean rating of 65.6 and a negative viewing-order trend.
All profile summaries are calculated from the features used to form the clusters.

Participant mean ratings had a between-participant CV of 0.16.
The between-participant CVs of the individual SD, CV, and range features were 0.81, 0.98, and 0.80, respectively.
Thus, the recruited sample varies in both overall rating level and the amount of rating variation across performances.
The split-half correlation of participant mean ratings was $r=0.866$, indicating high within-sequence rank-order consistency.

\subsection{B: Viewing-Order Associations in Ratings}
\paragraph{Method.}
For each of the 228 participants, we fitted a linear regression of performance rating on numeric viewing position, using that participant's observed ratings.
A two-sided one-sample $t$-test evaluated the mean of these participant slopes against zero.
Each participant contributes one slope to this test.
The coefficient is measured in rating points per viewing position under the randomized presentation protocol.

\paragraph{Results.}
The average individual slope was 0.488 points per position ($t(227)=5.42$, $p<0.001$).
The position-wise mean rating rose from 77.95 (SD 19.70) at the first position to 83.15 (SD 16.83) at the tenth, a difference of 5.20 points.
The endpoint difference summarizes the ratings available at positions 1 and 10; the mean slope uses all observed positions within each participant.

Ratings were higher at later viewing positions on average. Familiarity, mood, and rating calibration are possible explanations for this association.

\begin{figure}[t]
\centering
\includegraphics[width=\columnwidth]{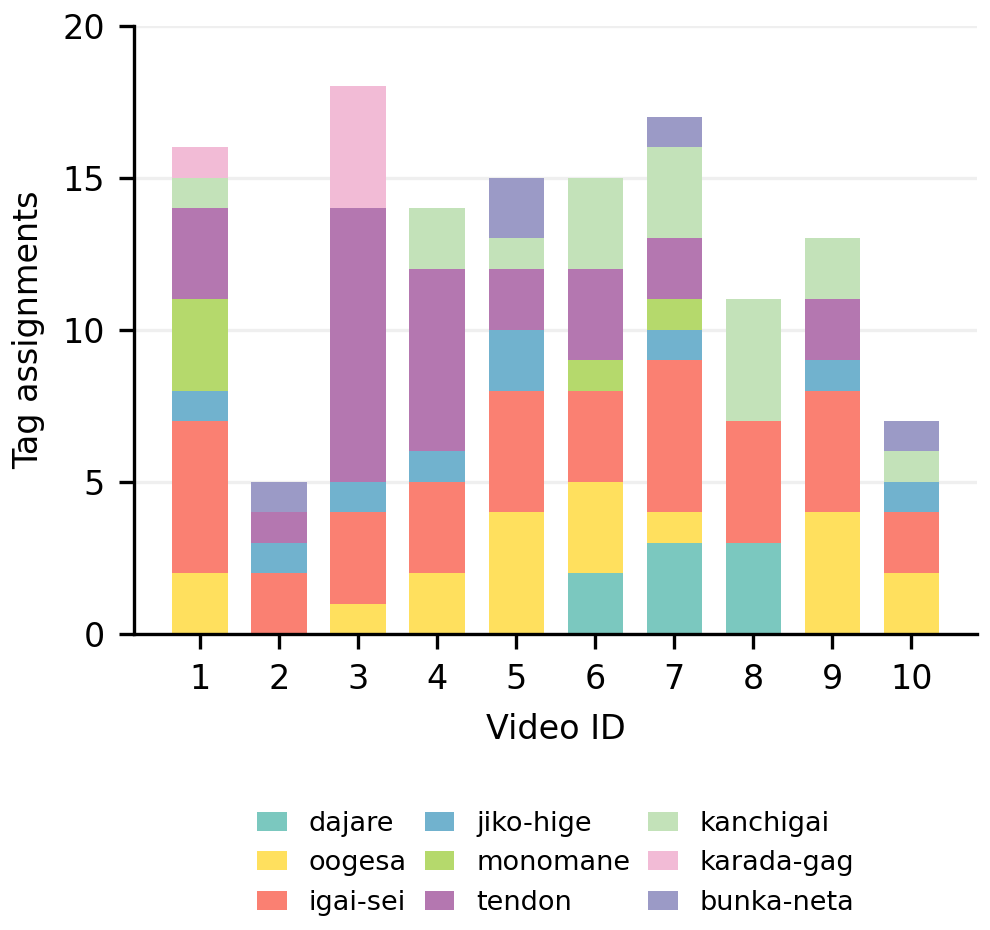}
\caption{Automated humor-tag assignments across the 10 performances. Each colored segment counts assignments of one tag. The bars sum to 131 assignments over 77 instances; multi-label instances contribute to multiple segments.}
\label{fig:tagdistribution}
\end{figure}
\begin{table}[t]
\centering\small
\setlength{\tabcolsep}{3pt}
\caption{Humor categories and audio-based response rates with 95\% bootstrap confidence intervals. The analysis includes 77 instances and 207 viewers. Responses are measured within 3 seconds of instance onset. Multi-label instances contribute to every assigned category. }
\label{tab:humor}
\begin{tabular}{lrrc}\toprule
Category & $n$ & Share & Response (\%)\\\midrule
Wordplay & 8 & 10.4\% & 46.7 (44.7--48.8)\\
Exaggeration & 19 & 24.7\% & 56.1 (53.4--58.8)\\
Unexpected twist & 35 & 45.5\% & 54.1 (51.8--56.4)\\
Self-deprecation & 9 & 11.7\% & 57.5 (53.9--61.4)\\
Impersonation & 5 & 6.5\% & 50.9 (48.4--52.7)\\
Repetition & 28 & 36.4\% & 53.5 (51.6--55.5)\\
Misunderstanding & 17 & 22.1\% & 51.4 (48.3--54.8)\\
Physical gag & 5 & 6.5\% & 50.7 (50.2--51.7)\\
Cultural reference & 5 & 6.5\% & 59.0 (53.8--64.3)\\
\bottomrule\end{tabular}
\end{table}
\begin{figure}[t]
\centering
\includegraphics[width=\columnwidth]{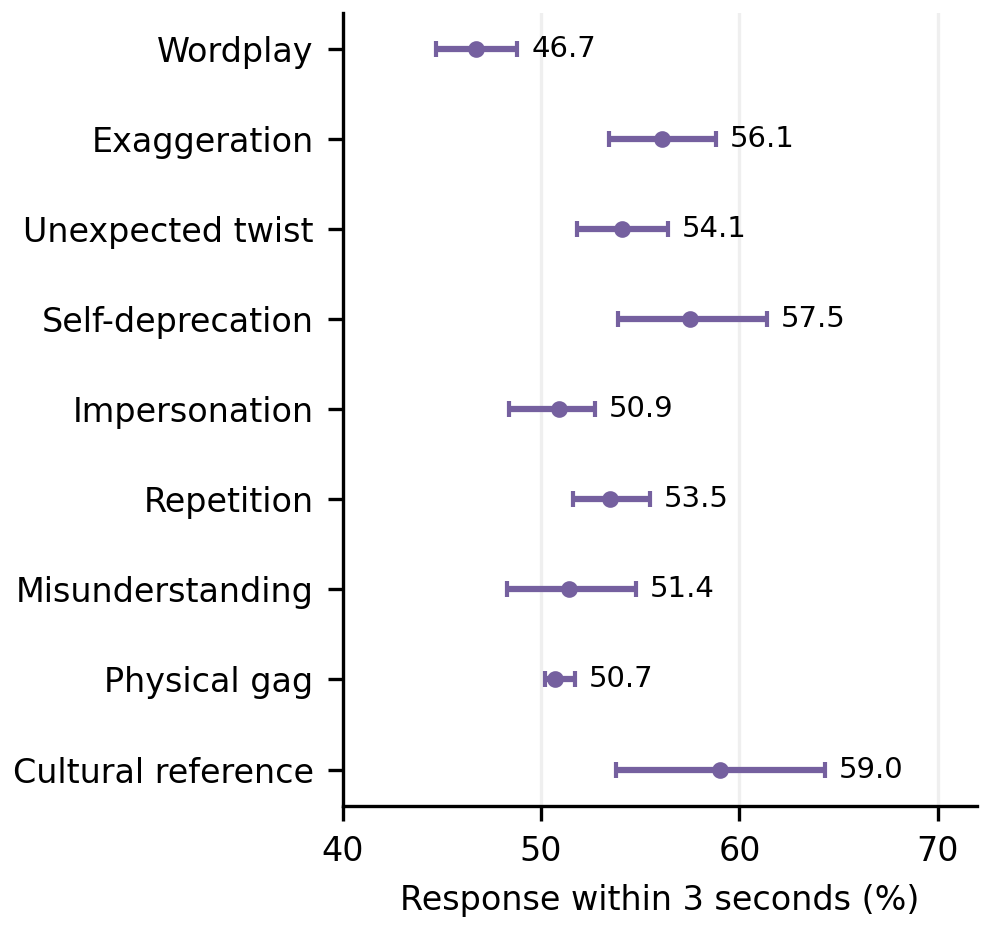}
\caption{Candidate vocal response rates within 3 seconds of humor-instance onset, by automated humor tag, among 207 viewers. Points and error bars show category-level response percentages and 95\% bootstrap confidence intervals.}
\label{fig:responserates}
\end{figure}
\subsection{C: Automated Humor Tags and Candidate Vocal Responses}
\paragraph{Method.}
Whisper~\cite{radford2022whisper} provided transcriptions of the 10 performances, and GPT-5-mini assigned humor tags to 77 stimulus instances.
The nine tags were wordplay, exaggeration, unexpected twist, self-deprecation, impersonation, repetition, misunderstanding, physical gag, and cultural reference.
Multiple tags could be assigned to one instance.

Candidate vocal responses were derived from microphone audio using frame-level root-mean-square energy, zero-crossing rate, spectral features, and peak detection.
We recorded whether a candidate vocal response occurred within 3 seconds of each humor-instance onset.
The complete-case sample contains 207 viewers with response data for the relevant instances.
We report the response percentage for each humor category.
Multiple tags allow an instance to contribute to more than one category summary.

To compare humor categories, we modeled viewer-by-instance responses using a marginal logistic model estimated with generalized estimating equations (GEE), participant-level clustering, an exchangeable working correlation, and video fixed effects. All humor-tag indicators entered the model jointly to accommodate multiple labels per instance. Let $y_{i,p}$ indicate whether viewer $p$ responded to instance $i$:
\begin{equation}
\operatorname{logit}\Pr(y_{i,p}=1)=\alpha+\sum_j\beta_j L_{ij}+\sum_v\gamma_v V_{iv},
\end{equation}
where $L_{ij}$ indicates humor tag $j$ and $V_{iv}$ indicates the performance containing instance $i$.
When GEE failed to converge, the fallback was an aggregated binomial generalized linear model with video fixed effects, HC3 robust covariance, and add-0.5 pseudo-counts for sparse or separated cells.
Pairwise differences between humor categories were evaluated with Wald tests and Benjamini--Hochberg false discovery rate (FDR) correction.

\paragraph{Results.}
The 77 instances received 131 tag assignments (Table~\ref{tab:humor}).
Unexpected twists were most frequent (35 instances, 45.5\%), followed by repetition (28, 36.4\%) and exaggeration (19, 24.7\%).
Category percentages use 77 instances as the denominator and total more than 100\% because of multiple assignments.
Figure~\ref{fig:tagdistribution} shows the assignments across the 10 performances.

After accounting for within-viewer correlation and performance differences, no statistically significant pairwise differences between humor categories were detected after FDR correction (all adjusted $p>0.05$).
Table~\ref{tab:humor} and Fig.~\ref{fig:responserates} show the descriptive response rates and 95\% bootstrap confidence intervals. Three-second response rates range from 46.7\% to 59.0\%. Cultural references have the highest descriptive rate (59.0\%, five instances), and wordplay has the lowest (46.7\%, eight instances). The category summaries are unadjusted; the pairwise tests account for performance and jointly included tags.

\section{Discussion and Limitations}
\subsection{Research Uses of the Collection Design}
The central contribution is the linkage among shared performances, repeated individual audiovisual recordings, performance ratings, and viewing order.
This organization supports comparisons of different viewers exposed to the same material and of the same viewer across multiple performances.
The rating profiles and within-sequence consistency support investigating prediction models that account for individual viewers.
Such evaluations can use held-out viewers and held-out performances to distinguish generalization across people from generalization across content.

Stimulus-linked facial and vocal features also motivate investigation of the timing and co-occurrence of visible and audible responses.
Performance-level ratings provide an overall evaluation that can be related to these temporally varying signals.
The positive viewing-order association suggests including presentation position in models of session data.

\subsection{Population, Stimulus, and Measurement Scope}
The sample includes 201 men among 241 participants and many frequent comedy viewers.
Recruitment, viewing familiarity, and completion can affect the observed distributions.
All participants were native Japanese speakers, and all 10 performances came from a single professional event.
Extending the corpus to other languages, audiences, and performance genres would enable cross-cultural comparisons.

Ratings are concentrated toward the upper end of the 0--100 scale, with a corpus mean of 82.0.
Ceiling constraints may contribute to the low variability of the high-rating profile.
SD, CV, and range share information, so their joint inclusion affects the clustering solution even after standardization.
The three profiles summarize the observed viewing sequence; their stability across separate occasions remains a question for longitudinal research.

Remote collection introduces variation in lighting, camera placement, microphones, and background sound.
Automated humor tags and audio-based response proxies introduce measurement uncertainty, particularly for visually grounded humor, quiet reactions, speech, and stimulus leakage.
Category comparisons involve 77 instances from 10 performances, with small samples for some categories and overlapping labels.

\section{Conclusion}
ManzaiSet links 10 shared professional manzai performances to 2,301 individual viewer--performance recordings from 241 native Japanese speakers, totaling 191.8 hours.
The collection combines facial video, microphone audio, performance-level ratings, and viewing-order information within a repeated-viewing design.
Exploratory analyses reveal differences in viewer rating patterns, consistency within the viewing sequence, and higher average ratings at later viewing positions. No statistically significant differences in viewer responses were found between humor categories after multiple-comparison correction.
The corpus supports research on how different viewers respond to shared Japanese comedy performances.

\section*{Acknowledgments}
This work was supported by JST Moonshot R\&D Grant JPMJMS2012.
We thank Yoshimoto Kogyo Holdings Co., Ltd. for providing access to professional performances and supporting data collection.
We also thank Hongil Yang, Shunsuke Noma, and Naoki Shibuya for their assistance with data acquisition.
{\small\bibliographystyle{ieeenat_fullname}\bibliography{main}}
\end{document}